\documentclass[10pt,twocolumn,letterpaper]{article}

\usepackage{iccv}              

\usepackage{epsfig,graphics} 
\usepackage{amsmath,amssymb} 
\usepackage{mathrsfs} 
\usepackage{mathtools} 
\usepackage{esdiff} 
\usepackage{nicefrac,xfrac} 
\usepackage{bm} 
\usepackage{xcolor} 
\usepackage{paralist} 
\usepackage[normalem]{ulem} 

\usepackage[ruled]{algorithm2e} 

\SetAlFnt{\small}
\SetAlCapFnt{\small}
\SetAlCapNameFnt{\small}
\SetAlCapHSkip{0pt}

\usepackage[accsupp]{axessibility}  

\newcommand{\email}[1]{\url{#1}}
\newcommand{\tss}[1]{\textsuperscript{#1}}

\newif\ifdraft
\draftfalse
\ifdraft

\newcommand{\revdel}[1]{{\color{red}\sout{#1}}}

\newcommand{\TODO}[1]{{\color{purple}TODO: {#1}}}

\else

\newcommand{\revdel}[1]{}

\newcommand{\TODO}[1]{}
\fi

\newif\ifarxiv
\arxivtrue

\definecolor{iccvblue}{rgb}{0.21,0.49,0.74}
\usepackage[pagebackref,breaklinks,colorlinks,allcolors=iccvblue]{hyperref}

\usepackage[capitalize]{cleveref}
\crefname{section}{Sec.}{Secs.}
\Crefname{section}{Section}{Sections}
\Crefname{table}{Table}{Tables}
\crefname{table}{Tab.}{Tabs.}

\usepackage{scrextend}
\deffootnote{0em}{0em}{\thefootnotemark}

\def\paperID{*****} 
\def\confName{ICCV}
\def\confYear{2025}

\title{VQualA 2025 Challenge on Face Image Quality Assessment: \\Methods and Results}

\author{
Sizhuo Ma\textsuperscript{*}
\and
Wei-Ting Chen\textsuperscript{*}
\and
Qiang Gao\textsuperscript{*}
\and
Jian Wang\textsuperscript{*}
\and
Chris Wei Zhou\textsuperscript{*}
\and
Wei Sun
\and
Weixia Zhang
\and
Linhan Cao
\and
Jun Jia
\and
Xiangyang Zhu
\and
Dandan Zhu
\and
Xiongkuo Min
\and
Guangtao Zhai
\and
Baoying Chen
\and
Xiongwei Xiao
\and
Jishen Zeng
\and
Wei Wu
\and
Tiexuan Lou
\and
Yuchen Tan
\and
Chunyi Song
\and
Zhiwei Xu
\and
MohammadAli Hamidi
\and
Hadi Amirpour
\and
Mingyin Bai
\and
Jiawang Du
\and
Zhenyu Jiang
\and
Zilong Lu
\and
Ziguan Cui
\and
Zongliang Gan
\and
Xinpeng Li
\and
Shiqi Jiang
\and
Chenhui Li
\and
Changbo Wang
\and
Weijun Yuan
\and
Zhan Li
\and
Yihang Chen
\and
Yifan Deng
\and
Ruting Deng
\and
Zhanglu Chen
\and
Boyang Yao
\and
Shuling Zheng
\and
Feng Zhang
\and
Zhiheng Fu
\and
Abhishek Joshi
\and
Aman Agarwal
\and
Rakhil Immidisetti
\and
Ajay Narasimha Mopidevi
\and
Vishwajeet Shukla
\and
Hao Yang
\and
Ruikun Zhang
\and
Liyuan Pan
\and
Kaixin Deng
\and
Hang Ouyang
\and
Fan Yang
\and
Zhizun Luo
\and
Zhuohang Shi
\and
Songning Lai
\and
Weilin Ruan
\and
Yutao Yue
}

\begin{document}
\maketitle

\renewcommand{\thefootnote}{}
\footnotetext{$^{*}$Sizhuo Ma (\email{sma@snap.com}), Wei-Ting Chen (\email{weitingchen@microsoft.com}),  Qiang Gao (\email{qgao@snap.com}), Jian Wang (\email{jwang4@snap.com}) and Chris Wei Zhou (\email{zhouw26@cardiff.ac.uk}) are the challenge organizers. The other authors are participants of the VQualA 2025 Challenge on Face Image Quality Assessment.}

\begin{abstract}
Face images play a crucial role in numerous applications; however, real-world conditions frequently introduce degradations such as noise, blur, and compression artifacts, affecting overall image quality and hindering subsequent tasks. To address this challenge, we organized the VQualA 2025 Challenge on Face Image Quality Assessment (FIQA) as part of the ICCV 2025 Workshops. Participants created lightweight and efficient models (limited to 0.5 GFLOPs and 5 million parameters) for the prediction of Mean Opinion Scores (MOS) on face images with arbitrary resolutions and realistic degradations. Submissions underwent comprehensive evaluations through correlation metrics on a dataset of in-the-wild face images. This challenge attracted 127 participants, with 1519 final submissions. This report summarizes the methodologies and findings for advancing the development of practical FIQA approaches.

\end{abstract}
    
\section{Introduction}
\label{sec:intro}
In recent years, face images have become integral to a wide variety of applications, including video communication, photography, augmented reality, and digital content creation. However, real-world face images are frequently captured under non-ideal conditions due to environmental constraints and hardware limitations, resulting in common degradations such as noise, blur, compression artifacts, and poor lighting. These degradations not only diminish perceived image quality but also negatively impact downstream image processing tasks like enhancement, editing, and synthesis. Moreover, compromised image quality can adversely affect the performance and generalization ability of data-driven models, including large-scale vision systems and generative models, which depend on high-quality face image datasets for effective training~\cite{karras2017progressive,ruiz2023dreambooth}. Thus, the development of robust generic FIQA methods capable of accurately quantifying perceptual degradation levels has become increasingly critical~\cite{su2023going,chen2024dsl}.

To advance research in this area, we introduce the VQualA 2025 Challenge on Face Image Quality Assessment, held in conjunction with the ICCV 2025 Workshops. This challenge focuses specifically on evaluating the perceptual quality of face images on an arbitrary scale affected by real-world degradations, emphasizing accuracy within stringent computational constraints. Participants are tasked with developing efficient and lightweight models capable of predicting the MOS of face images under conditions such as blur, noise, and low illumination. To reflect realistic deployment scenarios, submissions must adhere to computational constraints, including a maximum of 0.5 GFLOPs and fewer than 5 million parameters. Model performance will be rigorously evaluated using no-reference image quality metrics and extensive subjective human studies to ensure alignment with human perceptual judgments.

The primary objective of this challenge is to encourage innovation in efficient and precise FIQA models suitable for real-time deployment on mobile and edge devices, ultimately advancing the broader field of perceptual quality assessment and enabling practical, real-world applications.

This challenge garnered significant interest, attracting 127 registered participants. Throughout the development phase, participants submitted 1058 entries, followed by 461 submissions during the final testing phase. Ultimately, 13 teams successfully submitted their final models and accompanying fact sheets, each providing detailed methodologies for face image quality assessment. \cref{sec:methods} presents a comprehensive analysis and summary of submitted methods. We anticipate that this challenge will contribute meaningfully to the ongoing progress of face image quality assessment methods, particularly in real-world scenarios under computational constraints.

This challenge is one of several associated with the VQualA Workshop at ICCV 2025, including: Image Super-Resolution Generated Content Quality Assessment~\cite{isrgcq2025iccvw}, Visual Quality Comparison for Large Multimodal Models~\cite{zhu2025vqa}, GenAI-Bench AIGC Video Quality Assessment~\cite{genai-bench2025iccvw}, Engagement Prediction for Short Videos~\cite{li2025evqa} and Document Image Quality Assessment~\cite{diqa2025iccvw}.
\section{VQualA FIQA Challenge}
\label{sec:challenge}

\subsection{Datasets and Evaluation}
To ensure a fair evaluation of participant solutions, we curated distinct training, validation, and testing datasets for this challenge. Our training set comprises 27,686 images, and our validation set contains 1,000 images, all collected from CelebA~\cite{liu2015faceattributes} and Flickr. For the test set, we gathered 889 images exclusively from Flickr.

\paragraph{Variety in resolution.} A key challenge of this competition was developing a Face Image Quality Assessment (FIQA) method capable of handling in-the-wild images with diverse resolutions. Unlike previous datasets, such as GFIQA~\cite{su2023going}, our collected face images are not normalized and exhibit a wide range of resolutions, with short-edge dimensions varying from 224 to 1024 pixels. To generate labels for all datasets, we employed the state-of-the-art FIQA method, DSL-FIQA~\cite{chen2024dsl}. For each image, regardless of its resolution, 20 random patches were extracted and averaged to determine its quality score.

\paragraph{Evaluation.}
The challenge was structured into two distinct phases:
\begin{compactitem}
\item Development Phase: During this phase, participants were provided with the training images and their corresponding labels, along with the validation images. They were tasked with developing their solutions and uploading prediction results for the validation set, which were then compared against the ground truth.

\item Testing Phase: For the testing phase, participants were required to upload their model definitions and weights. The models then processed the unseen test images directly on our server, and the results were compared against the ground truth labels. We intentionally did not release the test dataset due to a strict constraint of 0.5 GFLOPs and 5M parameters. Releasing the test images could have led to participants using larger models to generate pseudo-labels for these images, subsequently training smaller models that overfit, thereby compromising the fairness of the competition. 

The awards were determined according to the testing phase scores. We use the average of SROCC and PLCC as the overall score:
\begin{equation}
    \mathrm{Score} = (\mathrm{SROCC} + \mathrm{PLCC}) / 2
\end{equation}
\end{compactitem}

\subsection{Baseline}
To facilitate the development of solutions, a MobileNetV2-based~\cite{sandler2018mobilenetv2} baseline was provided, which accepts $224 \times 224$ pixel image patches as input. During inference, multiple random patches were cropped from the original image and processed by the network. The resulting output scores were subsequently averaged to yield a final prediction. The baseline was trained using the Adam~\cite{kingma2015adam} optimizer with a learning rate of $5 \times 10^{-4}$ and a weight decay of $10^{-5}$. The loss function used was Mean Squared Error. Training was conducted for 20 epochs with a batch size of 64.

While satisfactory performance was achieved with an ensemble of 20 random crops, adherence to GFLOPs constraints necessitates the use of a single crop, which exhibits suboptimal performance (see \cref{tab:results}). This constraint poses a challenge for participants, requiring the development of optimal input-handling strategies, including appropriate resizing and cropping techniques, to maximize performance under computational limitations.

\subsection{Challenge Results}
\begin{table*}[t]
    \centering
    \caption{Challenge Results}
    \label{tab:results}
    \begin{tabular}{|c|l|c|c|c|c|c|}
        \hline
        \textbf{Rank} & \textbf{Team} & \textbf{Score} & \textbf{SROCC} & \textbf{PLCC} & \textbf{GFLOPS} & \textbf{NumParams[M]} \\
        \hline
        1 & ECNU-SJTU VQA Team & 0.9664 & 0.9692 & 0.9637 & 0.3313 & 1.1796 \\
        2 & MediaForensics & 0.9624 & 0.9624 & 0.9624 & 0.4687 & 1.5189 \\
        3 & Next & 0.9583 & 0.9630 & 0.9535 & 0.4533 & 1.2224 \\
        4 & ATHENAFace & 0.9566 & 0.9600 & 0.9533 & 0.4985 & 2.0916 \\
        5 & NJUPT-IQA-Group & 0.9547 & 0.9530 & 0.9564 & 0.4860 & 3.7171 \\
        6 & ECNU VIS Lab & 0.9406 & 0.9397 & 0.9415 & 0.4923 & 3.2805 \\
        7 & JNU620 & 0.9334 & 0.9413 & 0.9255 & 0.4097 & 3.2511 \\
        8 & ISeeCV & 0.9279 & 0.9282 & 0.9275 & 0.4890 & 0.9513 \\
        9 & RegNet & 0.9242 & 0.9262 & 0.9222 & 0.4895 & 4.0252 \\
        10 & Conquerit & 0.9038 & 0.9118 & 0.8958 & 0.2235 & 4.7795 \\
        11 & BIT\_ssvgg & 0.8727 & 0.8897 & 0.8557 & 0.5120 & 4.7242 \\
        12 & 2077Agent & 0.8432 & 0.8529 & 0.8335 & 0.2852 & 1.3005 \\
        13 & DERS & 0.6999 & 0.7098 & 0.6900 & 0.8980 & 6.0523 \\\hline
         & Baseline & 0.8309 & 0.8334 & 0.8283 & 0.3139 & 3.2511 \\
        \hline
    \end{tabular}
\end{table*}

\Cref{tab:results} summarizes the challenge results. A total of 13 teams submitted their solutions and accompanying fact sheets. 
The top-performing method attained a score of 0.9664, representing an improvement of over 0.13 relative to the baseline, notably with comparable computational complexity (GFLOPs) and a reduced number of parameters.

\ifarxiv
The subsequent section provides a detailed description of each submitted solution. A list of team members and affiliations are included in \cref{sec:team}.
\fi
\section{Teams and Methods} \label{sec:methods}

\subsection{Efficient Face Image Quality Assessment via Self-training and Knowledge Distillation (by ECNU-SJTU VQA Team)}
\begin{figure}[t]
  \centering
  \includegraphics[width=\linewidth]{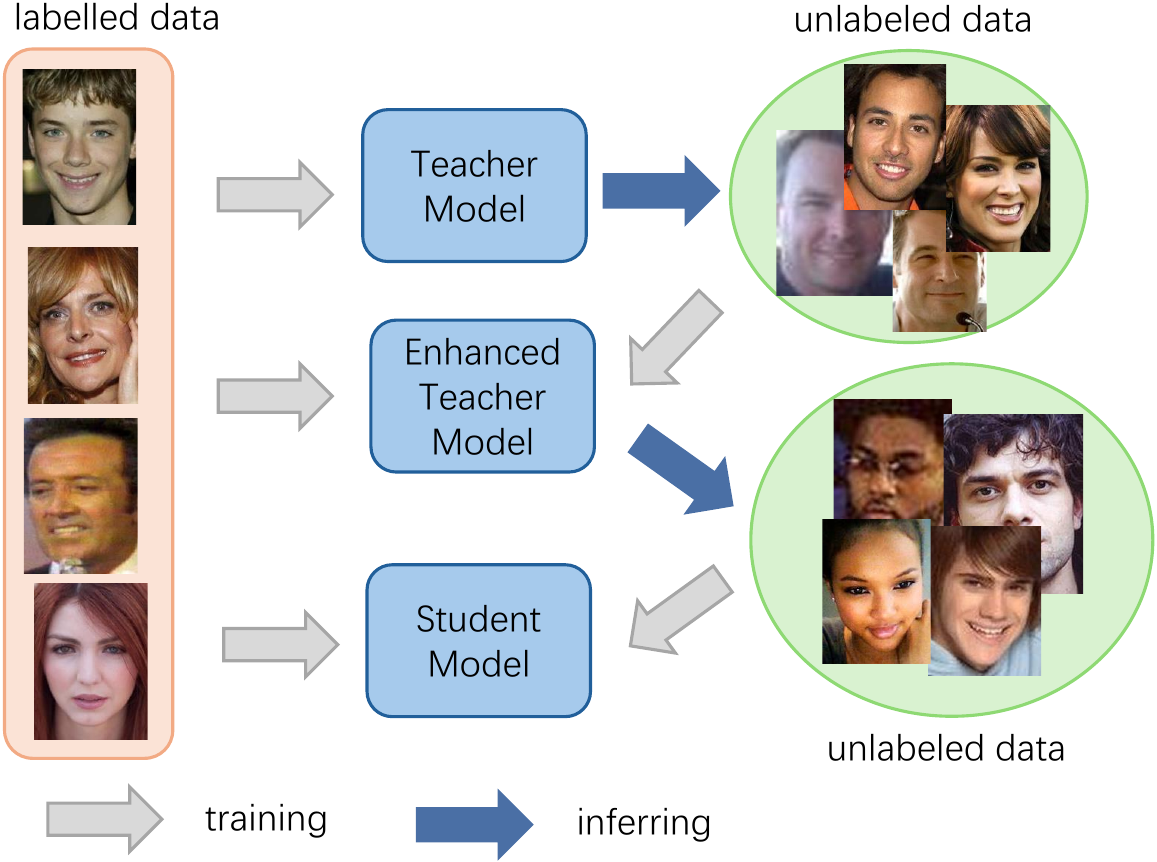}
  \caption{ECNU-SJTU VQA Team.}
  \label{fig:encu-sjtu-vqa}
\end{figure}

The \emph{ECNU-SJTU VQA Team} proposes a framework comprising two main stages, as described in \cref{fig:encu-sjtu-vqa}. First, they trained a teacher model using a self-training approach. Specifically, the Swin Transformer Base (Swin-B)~\cite{liu2021swin} was adopted. The classification head was removed and replaced with a two-layer multilayer perceptron (MLP) consisting of 128 and 1 hidden neurons, respectively, serving as the regression head. They began by training the teacher model on the labeled face image quality assessment (FIQA) dataset provided by the challenge organizers. Next, they collected a large-scale unlabeled face image dataset (approximately 200k images) from the Internet. The trained teacher model was then used to generate pseudo-labels for these images. They combined the labeled images and pseudo-labeled images to retrain the teacher model, thereby enhancing its performance through self-training. After this, the enhanced teacher was used to generate pseudo-labels for an additional set of collected face images (approximately 200k images) for the second-stage training.

In the second stage, they trained a student model using the labeled images, the first-round and the second-round pseudo-labeled images. The student model employed EdgeNeXt-XX-Small~\cite{maaz2022edgenext} as the backbone, with its classification head replaced by the same two-layer MLP regression head. Through learning from ground-truth data, the teacher-labeled data, and the enhanced teacher-labeled data, the student model achieved competitive performance, closely matching that of the enhanced teacher.

Their approach is inspired by the Self-training with Noisy Student framework~\cite{xie2020self}, but differs in two key aspects. First, unlike the original method which uses a larger model for iterative self-training, they retained the same architecture (i.e., Swin-B). Additionally, since their goal was to assess visual quality, they avoided introducing noise to the input images for self-training, as it might degrade their perceptual fidelity. Second, they further distilled the enhanced teacher into a lightweight student model to enable efficient image quality assessment. More details can be found in the challenge paper~\cite{sun2025efficient}.

\paragraph{Training details.} 
They implemented their framework using PyTorch 2.4~\cite{paszke2019pytorch}. For the two-round teacher training, they used the AdamW~\cite{loshchilov2019decoupled} optimizer with a learning rate of 1×10$^{-4}$, a weight decay of 1×10$^{-6}$, and a learning rate decay factor of 0.1 every 10 epochs. The model was trained for 30 epochs with a batch size of 32. They randomly selected 20\% of the provided labeled data as a validation set to determine the best-performing teacher model. The loss function was a combination of Mean Squared Error (MSE) loss and Pearson Linear Correlation Coefficient (PLCC) loss. During training, images were resized such that the small side was 448 pixels while preserving the aspect ratio, followed by a random crop of 448×448. For student training, the only difference was the image resolution: images were resized and cropped to 352×352 instead. The teacher model was trained using two NVIDIA RTX 3090 GPUs, while the student model was trained on a single RTX 3090.

\paragraph{Testing details.}
For testing, images were first resized so that the small side was 288 pixels while maintaining the aspect ratio, followed by a center crop of 288×288. The cropped images were then fed into the student network to derive the quality score.

\subsection{MSPT: A Lightweight Face Image Quality Assessment Method with Multi-Stage Progressive Training (by MediaForensics)}
\begin{figure*}[t]
  \centering
  \includegraphics[width=0.95\linewidth]{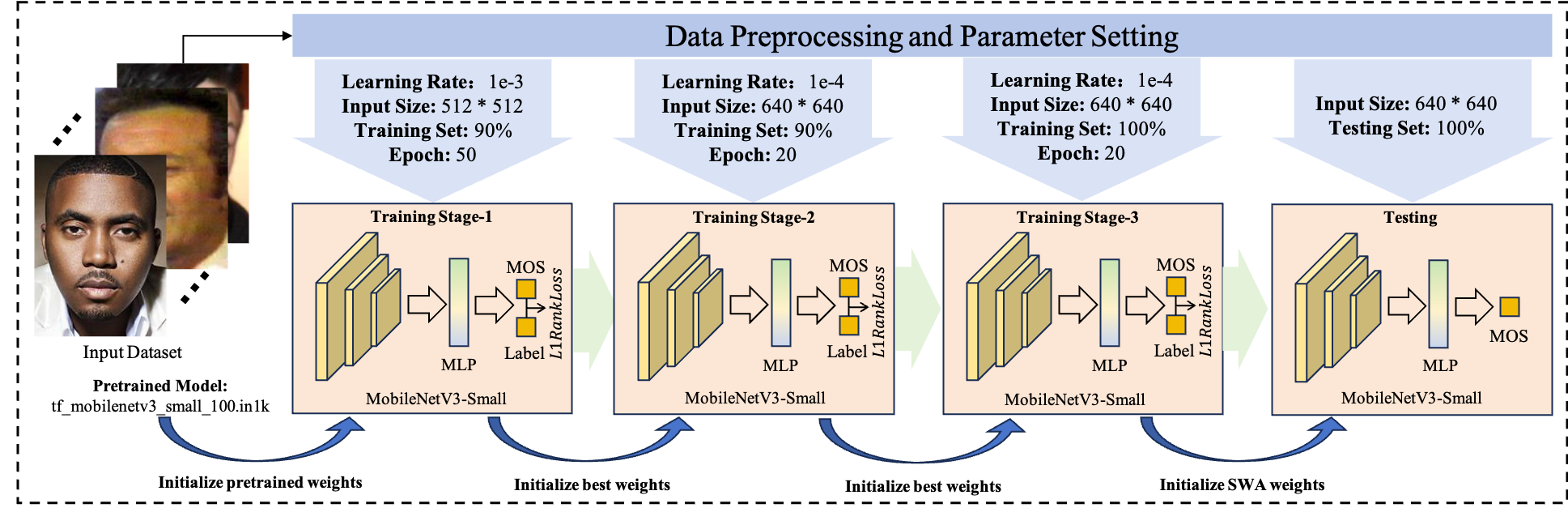}
  \caption{MediaForensics.}
  \label{fig:mediaforensics}
\end{figure*}

The \emph{MediaForensics} Team utilizes multi-stage progressive training (MSPT) for lightweight face image quality assessment, enhancing model performance through three progressive training stages~\cite{xiao2025msptlightweightfaceimage}, as shown in \cref{fig:mediaforensics}. In the first stage, 90\% of the training set was used to train the model at a size of 512 for 50 epochs. In the second stage, the size was increased to 640 for fine-tuning over 20 epochs. In the third stage, the entire training set was used again for fine-tuning at a size of 640 for another 20 epochs. Finally, Stochastic Weight Averaging (SWA)~\cite{izmailov2018averaging} was applied to the model weights from the third stage to improve generalization and assessment effectiveness.

\paragraph{Training details.}
They adopted the lightweight model "tf\_mobilenetv3\_small\_100.in1k"~\cite{howard2019searching}, loaded using the \emph{timm}~\cite{wightman2019timm} library. During training, they started with the first stage using 90\% of the training set to train the model for 50 epochs, with the remaining 10\% used as a validation set, employing random cropping to a size of 512 and a learning rate of 0.001. In the second stage, they increased the size to 640 and continued fine-tuning for 20 epochs, reducing the learning rate to 0.0001. Finally, in the third stage, they maintained the size at 640 and the learning rate at 0.0001 to fine-tune the model on the entire training set for another 20 epochs. Throughout the three training stages, they used the L1RankLoss~\cite{wen2021strong} as the loss function, AdamW~\cite{loshchilov2019decoupled} as the optimizer, and a CosineAnnealingLR~\cite{loshchilov2016SGDRSG} scheduler with a period of 5, with a batch size of 256. Training was conducted on 4 NVIDIA 1080Ti GPUs, utilizing AMP (Automatic Mixed Precision) for mixed precision training. The data augmentations they applied included HorizontalFlip, RandomRotate90, PadIfNeeded, and RandomResizedCrop. To obtain the final inference model, they applied Stochastic Weight Averaging (SWA) to all the weights saved from the third stage to derive the final inference model weights.

\paragraph{Testing details.}
During the testing phase, they used the SWA model weights obtained from the third stage for inference. The image preprocessing included PadIfNeeded and CenterCrop, with an inference size of 640.

\subsection{Towards Robust No-Reference Image Quality Assessment via Prompt-Aware Alignment and Multi-Level Distillation (by Next)}
\begin{figure}[t]
  \centering
  \includegraphics[width=0.9\linewidth]{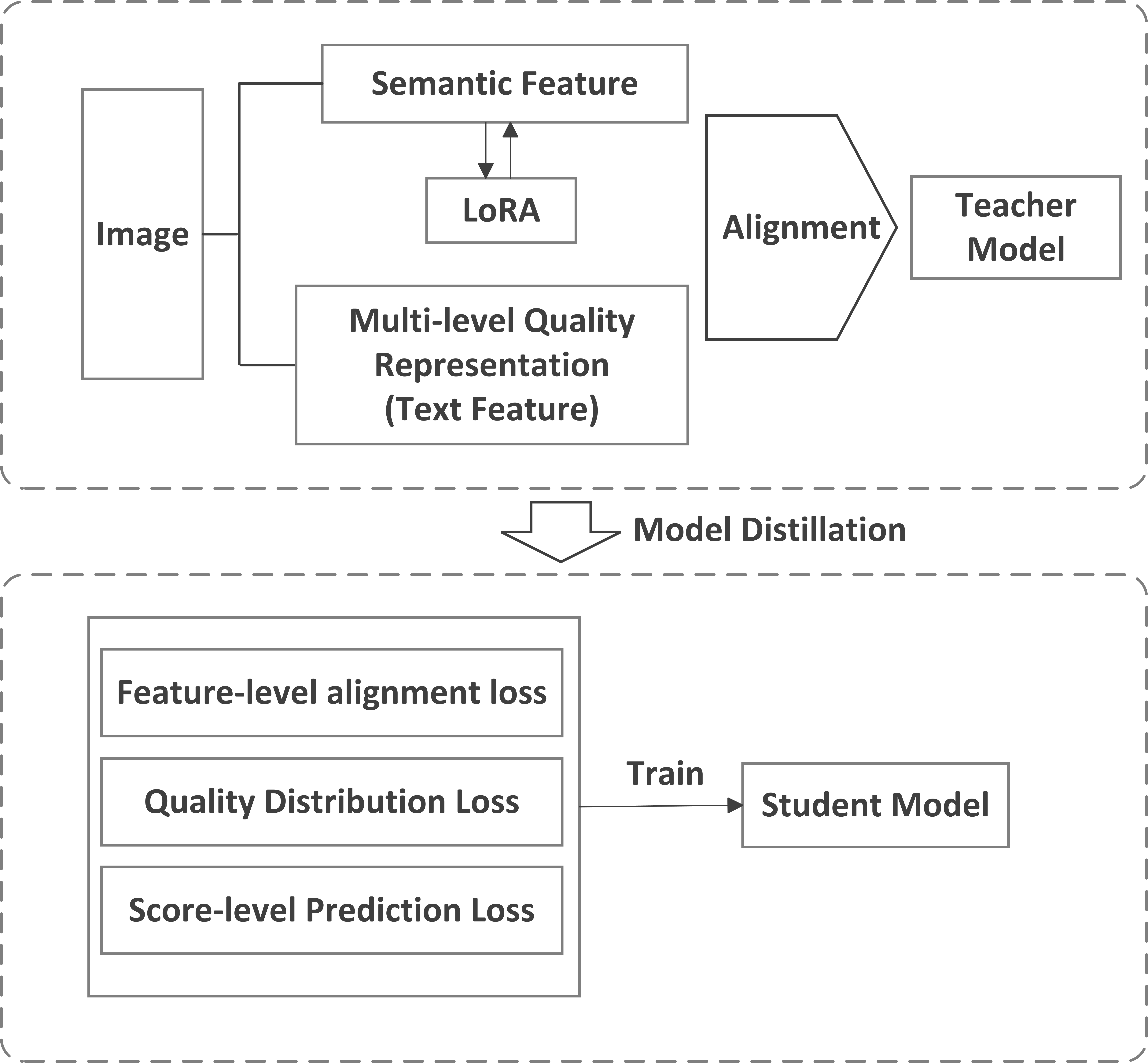}
  \caption{Next.}
  \label{fig:next}
\end{figure}

The \emph{Next} Team adopted the CLIP~\cite{radford2021learning} pre-trained vision-language model as the backbone for no-reference image quality assessment (NR-IQA). Instead of full fine-tuning, which requires updating all parameters, they introduced a multi-level LoRA\cite{hu2022lora}-based lightweight tuning strategy to adapt the general knowledge in CLIP to quality-related tasks with minimal training cost.

To enhance perceptual quality modeling, they designed a multi-level quality-aware prompt learning mechanism to predict the distribution over image quality grades. Specifically, each quality level is associated with a dedicated group of learnable prompt tokens, which are embedded using the CLIP text encoder. These prompts serve as semantic anchors to guide the alignment between visual features and discrete quality levels.

To further distill the knowledge into a compact student model, they constructed a hierarchical quality-aware loss framework that operates on three levels: feature-level alignment, quality score distribution matching, and alignment between predicted scores and ground-truth labels. This multi-level supervision enabled efficient and effective knowledge distillation, allowing the student model to retain both low-level distortion cues and high-level semantic consistency with minimal complexity.

\paragraph{Training details.}
Their training process followed a two-stage framework: teacher model training followed by student model distillation, as demonstrated in \cref{fig:next}. The teacher model was based on CLIP's vision backbone (ViT-B/32), while the student model adopted a lightweight MobileNetV3-Small~\cite{howard2019searching} architecture for efficiency.
They first pre-trained the teacher model on the GFIQA-20k dataset, which provides high-quality annotations for generic image quality assessment in similar domains. In the second stage, they fine-tuned the model on the target dataset’s training split to adapt to the specific task. The student model was then trained via knowledge distillation from the fine-tuned teacher model using their proposed multi-level supervision strategy.
Training was conducted on a single NVIDIA GPU. The optimizer used was AdamW~\cite{loshchilov2019decoupled}, with a learning rate of 5e-4 and a weight decay of 1e-2. For learning rate scheduling, they employed a two-phase schedule: a linear warm-up over the first 3 epochs (using LinearLR), followed by cosine annealing for the remaining epochs (using CosineAnnealingLR). These were implemented via SequentialLR, with a milestone at the end of warm-up.
In total, the full training process (including teacher training and student distillation) took approximately 18 hours. This two-stage strategy allowed the model to first capture high-level semantic quality cues and then efficiently transfer them into a compact model optimized for deployment.

\paragraph{Testing Details.}
During testing, each input image was first resized by scaling its shorter side to 620 pixels while preserving the original aspect ratio. Then, a center crop of size 620 × 620 was applied to obtain a square region. The cropped image was subsequently normalized using the mean and standard deviation consistent with the CLIP preprocessing pipeline. The preprocessed image was then fed into the trained model to predict the quality score.

\subsection{Face Image Quality Assessment via \\ Lightweight Ensemble Learning and \\ Correlation-Driven Optimization\\ (by ATHENAFace)}
\begin{figure}[t]
  \centering
  \includegraphics[width=\linewidth]{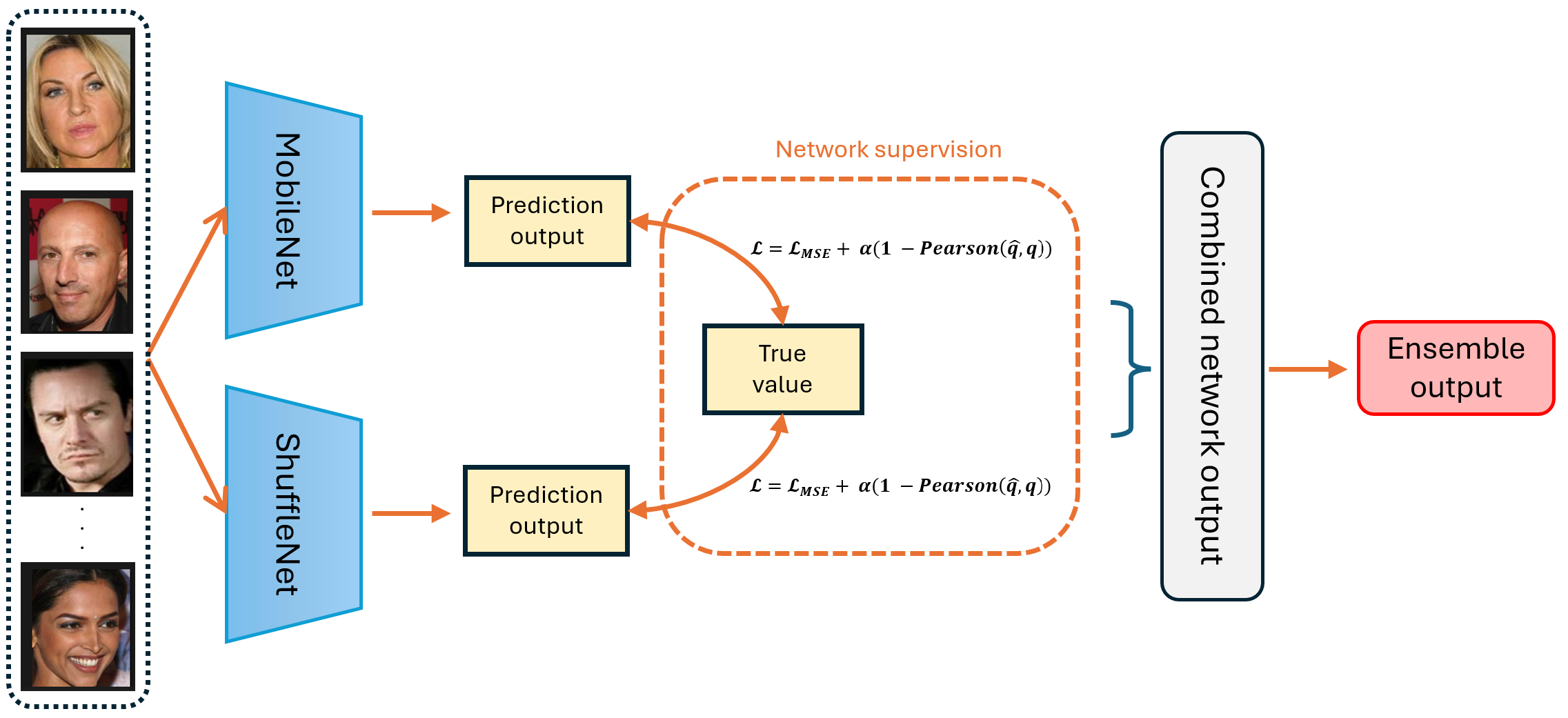}
  \caption{ATHENAFace.}
  \label{fig:athenaface}
\end{figure}

The \emph{ATHENAFace} Team proposed a two-branch ensemble method for image quality assessment, as shown in \cref{fig:athenaface}. Their approach combines MobileNetV3-Small~\cite{howard2019searching} and ShuffleNetV2~\cite{ma2018shufflenet}-based regression models initialized by training on the ImageNet~\cite{krizhevsky2012imagenet} dataset. An end-to-end training on a merged dataset of the competition's original data and the GFIQA-20k dataset was then performed. Subsequently, both models were fine-tuned solely on the competition's original dataset using progressive unfreezing: initially freezing the backbone and later unfreezing it to refine learned representations. The final prediction was made by averaging the outputs of the two models. During training, the models' predictions were supervised using the correlation-aware loss, which combines Mean Squared Error (MSE) with a Pearson correlation loss. This encouraged not only accurate predictions but also alignment with the relative ranking of perceptual quality scores, which is crucial in perceptual quality tasks. The used loss function is shown in the following:
\begin{equation}
\mathcal{L} = \mathcal{L}_{MSE}(q_i, \hat{q}_i) + \alpha \mathcal{L}_{Corr}(q_i, \hat{q}_i).
\end{equation}
The MSE component is defined as:
\begin{equation}
    \mathcal{L}_{\text{MSE}} = \frac{1}{N} \sum_{i=1}^N \left( q_i - \hat{q}_i \right)^2,
\end{equation} 
and, the correlation loss is defined as:
\begin{equation}
    \mathcal{L}_{\text{Corr}} = 1 - \text{Pearson}(q_i, \hat{q}_i).
\end{equation}
More details can be found in the challenge paper~\cite{hamidi2025alightweight}.

\paragraph{Training details.}
Their method was implemented using PyTorch 2.5.1 and CUDA 12.1. They utilized MobileNetV3-Small and ShuffleNetV2 as base model architectures, trained with an Adam optimizer (weight\_decay: 1e-5, batch size: 64) and a StepLR scheduler (step\_size=5, gamma=0.1). A combined MSE + Pearson loss (MSECorrLoss) was employed. The training comprised 80 epochs in total, split between pretraining on combined data (original + GFIQA) for 20 epochs per model (at 5e-4 learning rate) and fine-tuning on original data only for another 20 epochs per model (at 1e-4 learning rate with 0.5$\times$ scaling), incorporating progressive unfreezing of the backbone at epoch 8. Training was conducted on an NVIDIA RTX 6000 Ada Generation 48 GB GPU, taking approximately 50 seconds per epoch for a total of about 1 hour. Efficiency optimization was achieved through the use of lightweight backbones and early stopping based on validation SRCC/PLCC.

\paragraph{Testing details.}
At inference time, they ensembled the predictions of both models by averaging their outputs. Input images are resized to $600\times 416$.

\subsection{Lightweight Spatial-Frequency Fusion Network for Blind Face Image Quality Assessment (by NJUPT-IQA-Group)}

\begin{figure}[t]
  \centering
  \includegraphics[width=\linewidth]{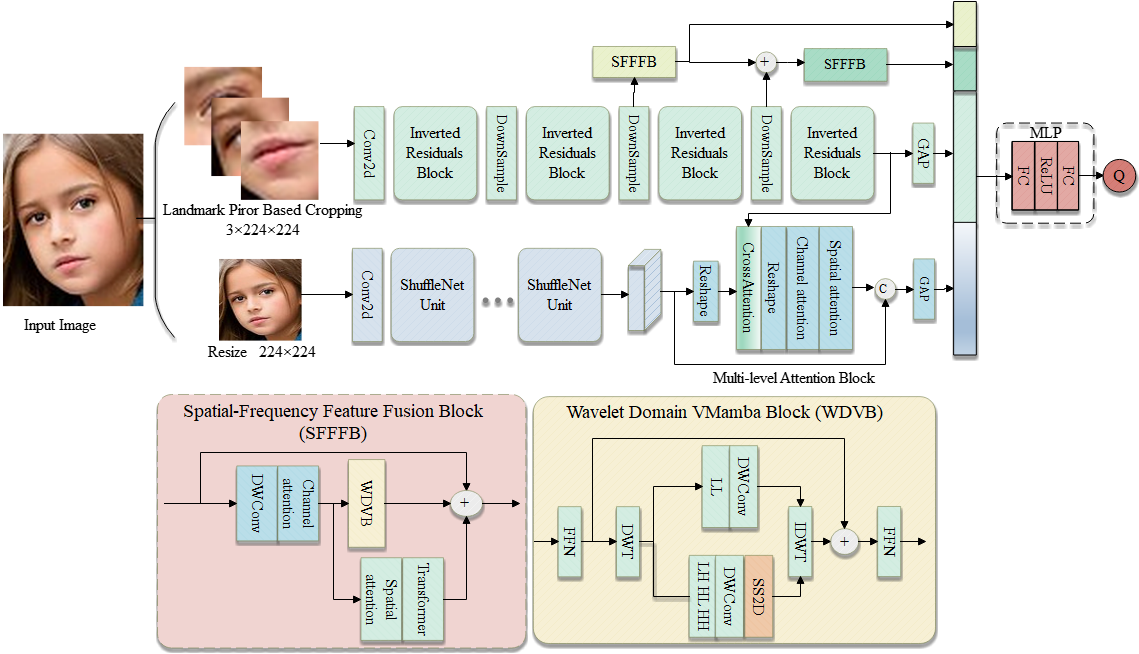}
  \caption{NJUPT-IQA-Group.}
  \label{fig:njupt}
\end{figure}

The \emph{NJUPT-IQA-GROUP} proposed a lightweight spatial-frequency feature fusion network for blind face image quality assessment (SFIQA). The model integrates wavelet transform and visual state space modeling to effectively capture spatial and frequency domain characteristics of facial images, as shown in \cref{fig:njupt}. SFIQA adopts a dual-branch architecture: a spatial-local feature extraction branch and a spatial-global feature extraction branch. 

In the local branch, MobileNetV3~\cite{howard2019searching} served as the backbone network for efficient local feature encoding. To enhance multi-level feature fusion, they introduced a novel spatial-frequency feature fusion block (SFFFB), which aggregates features from the last two layers of MobileNetV3. At the core of SFFFB lies the wavelet domain VMamba~\cite{liu2024vmamba} block (WDVB), which employs discrete wavelet transform (DWT) to decompose feature maps into one low-frequency (LL) and three high-frequency (LH, HL, HH) components. These high-frequency components, crucial for capturing fine-grained facial details such as edges, textures, and noise, were selectively enhanced via a 2D selective scan (SS2D) module. The refined features were then reconstructed using inverse discrete wavelet transform (IDWT), facilitating effective frequency-domain integration.
To further enrich the representation capacity, SFFFB incorporated channel attention, spatial attention, and a Transformer module, enabling the model to capture long-range dependencies and multi-scale contextual information.

In the global branch, they leveraged ShuffleNetV2-x0.5~\cite{ma2018shufflenet} to extract lightweight yet expressive global spatial features. To bridge the local and global feature representations, a Multi-level Attention Block was designed to facilitate effective interaction and integration between the two branches. This fusion strategy not only enhanced robustness but also significantly improved the discriminative power of the learned features.

Finally, the aggregated features from both branches were concatenated and passed through a multi-layer perceptron (MLP) to produce the final face image quality score.

\paragraph{Training details.}
To effectively extract fine-grained local features, they proposed a Landmark Prior based Cropping strategy based on facial keypoint distribution. Specifically, this method selected three 224$\times$224
local patches from each face image through the following procedure:
First, approximately 15,000 randomly sampled images from the training dataset were processed using a lightweight facial landmark detection network, PFLD~\cite{guo2019pfld}, to extract facial keypoints. These keypoints were then used to construct a 600$\times$800 keypoint integral heatmap, which highlighted the most discriminative facial regions. Based on this integral map, three key regions—typically centered around the eyes and mouth—were cropped as local patches and used as inputs to the local feature extraction branch. Meanwhile, the globally scaled full-face image was fed into the global feature extraction branch.\
The proposed SFIQA model was trained on a single Nvidia RTX 3090 GPU with a batch size of 64 for 30 iterations. The initial learning rate was set to 1$\times10^{-4}$, with a decay factor of 0.9 applied every two epochs. The AdamW optimizer was used for optimization.


\subsection{MobileNetV4 in FIQA (by ECNU VIS Lab)}
The \emph{ECNU VIS Lab} Team built an MobileNetV4-driven method for FIQA on edge devices.

\paragraph{Training Details.} During the training phase, they built their method upon MobileNetV4~\cite{qin2024mobilenetv4}. They used the Adam optimizer with an initial learning rate of 1e-4 and applied cosine annealing for dynamic learning rate adjustment. Training was conducted on 8 H100 GPUs. In addition to the competition dataset, they incorporated external datasets GFIQA~\cite{su2023going} and CGFIQA~\cite{chen2024dsl} to enhance model performance. They also evaluated the model under different input resolutions and found that 512$\times$512 yielded the best results. However, increasing the input size led to a sharp rise in computational cost. To address this, they modified the stride of the convolutional layers in MobileNetV4 to ensure the model met the competition’s efficiency constraints. They found that initializing the model with pre-trained MobileNetV4 weights significantly improved its performance.

\paragraph{Testing Details.} During the testing phase, the input images were resized to 512$\times$512 and directly passed through the model for prediction.

\subsection{Progressive Learning Strategy for Face Image Quality Assessment (by JNU620)}
\begin{figure}[t]
  \centering
  \includegraphics[width=\linewidth]{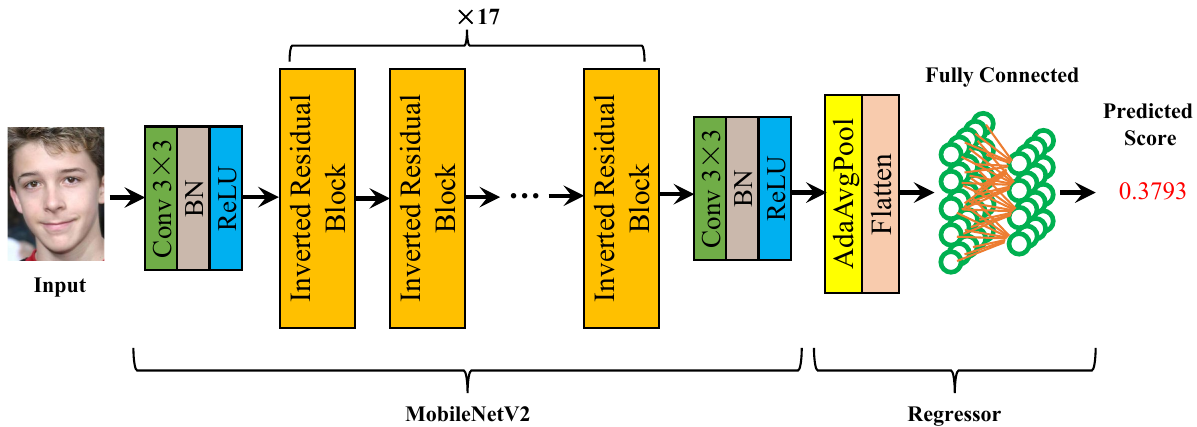}
  \caption{JNU620.}
  \label{fig:jnu620}
\end{figure}

A diagram of their network structure proposed by the \emph{JNU620} Team is illustrated in \cref{fig:jnu620}, which includes two parts: a MobileNetV2~\cite{sandler2018mobilenetv2} backbone for more effective and powerful feature representation and a regressor for mapping the quality-aware features to quality score spaces. They then introduced a progressive learning strategy to train the proposed model in two stages.

\paragraph{Training Details.}
The model was implemented in PyTorch. The estimated training time was 6 hours using an Nvidia Tesla V100 GPU (32 GB). The details of the training steps are as follows:

In the first stage, the model was trained on the in-the-wild training dataset provided by the organizer. The original face image was first resized to 244$\times$244, then randomly cropped into 224$\times$224 and flipped for augmentation. The mini-batch size was set to 32. The model was trained by minimizing the MSE loss function using the Adam optimizer~\cite{kingma2015adam}. The initial learning rate was set to 5e-4 and was reduced to 0.1 times the original value every 5 epochs. The total number of epochs was 50.

In the second stage, the model was trained again on 30k in-the-wild face images. The original face image was first resized to 276$\times$276, then randomly cropped into 256$\times$256 and flipped for augmentation. The mini-batch size was set to 32. The model was fine-tuned by minimizing the MSE loss function. The initial learning rate was set to 5e-5 and was reduced to 0.1 times the original value every 5 epochs. The total number of epochs was 100.

\paragraph{Testing Details.}
During the testing phase, the face image was first resized to 276$\times$276, then center-cropped into 256$\times$256 for inference.

\subsection{RankCORE: Ranking-Aware Correlation Optimized Regression Estimator (by ISeeCV)}
\begin{figure*}[t]
  \centering
  \includegraphics[width=0.9\linewidth]{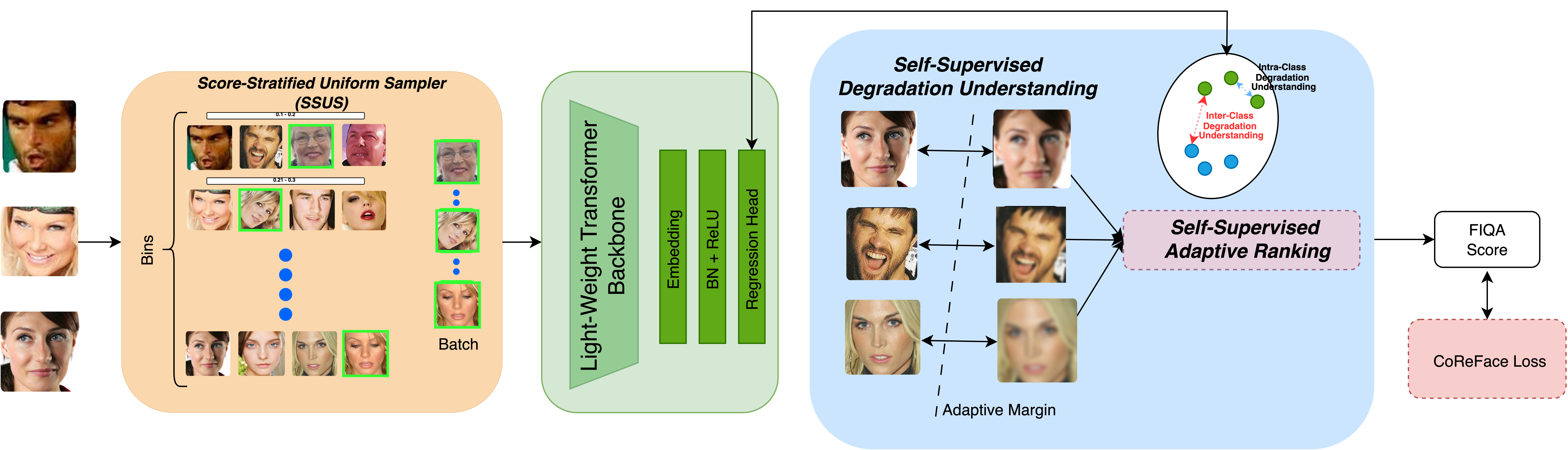}
  \caption{ISeeCV.}
  \label{fig:iseecv}
\end{figure*}

The \emph{ISeeCV} Team proposed a unified framework for face image quality assessment that jointly addresses both data sampling biases and prediction consistency across the score spectrum.
First, their Self-Supervised Adaptive Ranking loss (SSAR) encouraged the model to distinguish subtle degradations in a self-supervised manner.
Second, the Score-Stratified Uniform Sampler (SSUS) ensured each score band contributed equally to training, preventing overfitting to dense regions of the distribution and promoting gradient diversity. Finally, they optimized with a hybrid Wing-PLCC based CoReFace loss, which penalized both local prediction errors (via WingLoss) and global rank-order discrepancies (via a PLCC term), leading to more robust and well-calibrated quality estimates. These components formed a cohesive pipeline that improved both the accuracy and stability of face image quality predictions, as summarized in \cref{fig:iseecv}. 
\ifarxiv
More details are given in \cref{sec:rankcore} and the challenge paper~\cite{joshi2025rankcore}.
\fi

\paragraph{Training Details.}
Their method was implemented using the PyTorch 2.1.0 + cu118 deep learning framework. They employed the Adam optimizer with an initial learning rate of 1e-3, applying a MultiStepLR scheduler with gamma = 0.1 at defined epoch intervals. Training was conducted on an RTX A5000 GPU using only the dataset provided by the challenge, at resolution of $300\times 300$. The total training time was approximately 3 hours.

\paragraph{Testing Details.} They resize the images to $320\times 320$ and take a random crop of $300\times300$ for inference.

\subsection{Face Image Quality Assessment in the Wild with RegNet (by RegNet)}
\begin{figure}[t]
  \centering
  \includegraphics[width=0.7\linewidth]{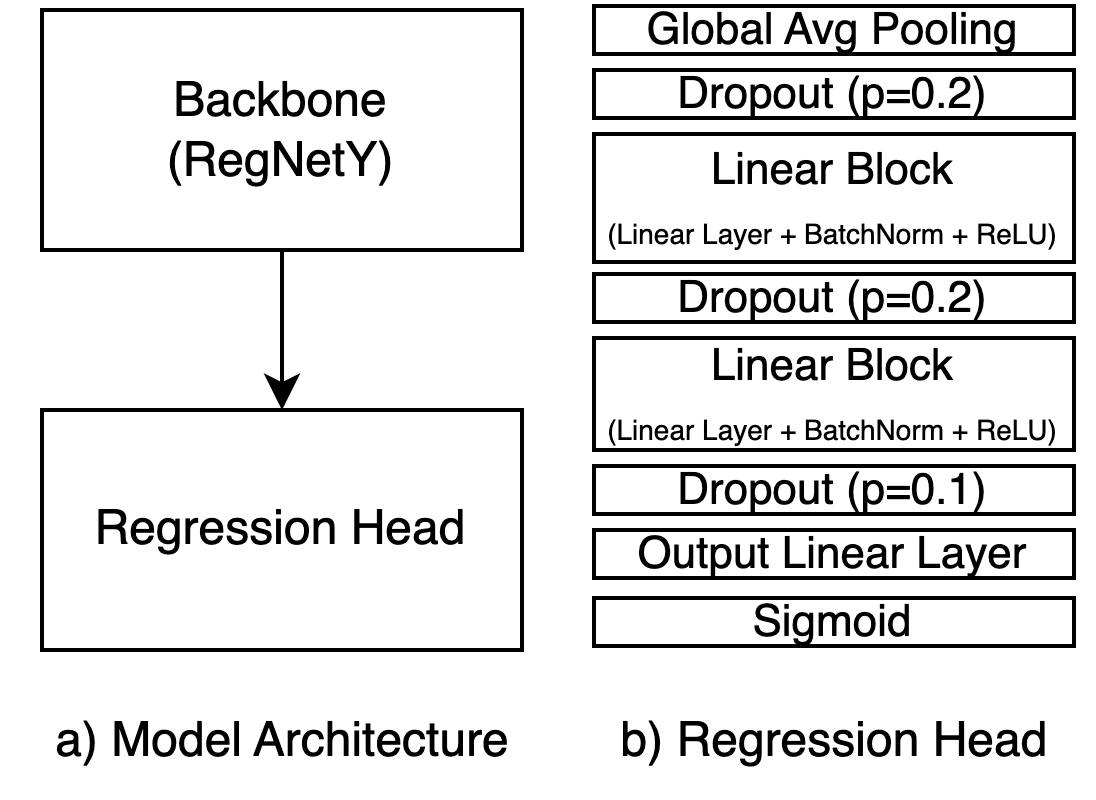}
  \caption{RegNet.}
  \label{fig:regnet}
\end{figure}
They addressed the challenge of perceptual face image quality assessment (FIQA) in a semi-wild scenario where the image resolution varied significantly, using a RegNet backbone~\cite{regnet}. The proposed method leveraged a RegNet model pre-trained on ImageNet1k as a feature extractor, on top of which a custom head composed of multiple fully connected layers and dropout was added. The model predicted continuous quality scores for face images, directly optimizing both mean squared error (MSE) and Pearson linear correlation coefficient (PLCC)~\cite{liu2023ntire} losses. The pipeline was designed for efficiency and generalization, handling diverse face image qualities and distributions.

\paragraph{Training details.} 
Training was conducted using the PyTorch framework. The RegNet backbone (regnet\_y\_400mf), initialized with ImageNet1k-pretrained weights, was extended with a head consisting of 3 fully connected layers with BatchNormalization, ReLU activation and interleaved with dropout for regularization. The model was optimized using a combination of MSE and PLCC losses, as this joint objective has been shown to improve correlation-based evaluation metrics for FIQA. The Adam optimizer was used with an initial learning rate of $5 \times 10^{-4}$, and learning rate scheduling was performed with StepLR (step size 5). Early stopping with a patience of 7 epochs was employed to prevent overfitting. All face images were resized to $240 \times 240$ pixels to standardize input and maintain computational efficiency, to meet the compute constraints of the challenge. Random rotation and random horizontal flip augmentations were applied during training to improve robustness. To address label imbalance, images with quality scores less than 0.1 or greater than 0.9 were oversampled by a factor of two. The training set consisted of both the main challenge dataset and the Comprehensive Generic Face IQA (CGFIQA) dataset~\cite{chen2024dsl}, with 30\% of the training data randomly selected as the validation set. All experiments were performed on a single NVIDIA GTX 1650 GPU. Throughout training, efficiency was prioritized by using a lightweight RegNet backbone, fixed input resolution, dropout, and early stopping.

\paragraph{Testing details.}
During testing, a similar image preprocessing pipeline as in training was applied, which includes resizing to $240 \times 240$ pixels and normalization. The trained model was evaluated on the validation and test sets using SRCC and PLCC as the primary metrics, following standard FIQA evaluation protocols. No test-time augmentation was used; predictions were generated from a single processed image per sample. The model’s parameter count and FLOPs were profiled to ensure compliance with efficiency constraints.

\ifarxiv
\subsection{MobileNetV3-Based FIQA (by Conquerit)}
The \emph{Conquerit} Team used a MobileNetV3-Large~\cite{howard2019searching} architecture. To evaluate and guide the training of their quality regression model, they adopt a correlation-based loss that encourages both value accuracy and ranking consistency between predicted scores $\hat{y}$ and ground-truth quality scores $y\in [0, 1]$.
\begin{equation}
    \mathcal{L}_{\text{corr}} = \alpha\cdot\mathcal{L}_{\text{PLCC}}+(1-\alpha)\cdot\mathcal{L}_{\text{Rank}},
\end{equation}
where $\alpha\in[0, 1]$ controls the balance between the two losses. The Pairwise Rank Loss approximates SRCC:
\begin{equation}
    \mathcal{L}_{\text{Rank}}=E_{ij}[\log(1+\exp(-(\hat{y_i}-\hat{y_j})\cdot\text{sign}(y_i-y_j)))].
\end{equation}

To enhance robustness against label noise and focus learning on harder examples, they propose a modified regression loss that combines label smoothing with a focal weighting scheme. This formulation, referred to as Focal Label Smoothing Loss, builds on the intuition of focal loss while adapting it to continuous regression targets. Given predicted score $\hat{y}$ and ground truth score $y$, they first apply label smoothing:
\begin{equation}
    \tilde{y} = y + \epsilon\cdot\mathcal{N}(0,1),
\end{equation}
where $\epsilon$ is the smoothing strength. This helps regularize overconfident predictions and improves generalization.
Next, they compute the squared error between predictions and smoothed targets:
\begin{equation}
    \text{MSE} = (\hat{y}-\tilde{y})^2.
\end{equation}
They then apply a \emph{focal weighting} to emphasize difficult samples:
\begin{equation}
    w_{\text{focal}}=(1-e^{-\text{MSE}})^\gamma, 
\end{equation}
where $\gamma$ is the focusing parameter that increases the loss contribution of harder examples (\ie larger errors). The loss is computed as
\begin{equation}
    \mathcal{L}_{\text{focal-smooth}}=s \cdot w_{\text{focal}}\cdot \text{MSE},
\end{equation}
where $s$ is an optional scaling factor for numerical stability or emphasis tuning. 

The final training objective balances the correlation loss and the focal label smoothing loss using another weighting parameter $\lambda$:
\begin{equation}
\mathcal{L}_\text{total}=\lambda\cdot\mathcal{L}_\text{corr} + (1-\lambda)\cdot\mathcal{L}_\text{focal-smooth}.
\end{equation}

\paragraph{Training details.}
They use ReduceLROnPlateau learning rate annealing during training. They use input size of (224,224,3) with normalization to (-1,1).

\paragraph{Testing details.}
Input images are resized to (240,240,3), center-cropped to (224,224,3) and normalized to (-1,1).

\subsection{Dual-Branch Network with Local and Global Perception for Face Image Quality Assessment (by BIT\_ssvgg)}
\begin{figure}[t]
  \centering
  \includegraphics[width=\linewidth]{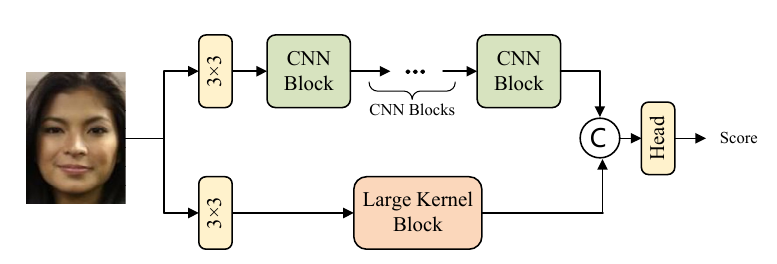}
  \caption{BIT\_ssvgg.}
  \label{fig:bit-ssvgg}
\end{figure}

The \emph{BIT\_ssvgg} Team observed that, to effectively assess the perceptual quality of facial images, it was crucial to capture both fine-grained local distortions and holistic structural degradations. Motivated by this, they adopted a dual-branch network design that leveraged complementary inductive biases, as shown in \cref{fig:bit-ssvgg}. Specifically, one branch was based on MobileNetV2~\cite{sandler2018mobilenetv2}, which captured localized texture variations and high-frequency artifacts through its hierarchical depthwise convolutional design and small receptive fields. However, while efficient, MobileNetV2 tended to lack global context aggregation, which is essential for assessing structure-aware degradations such as defocus blur, over-smoothing, or global compression artifacts.

To compensate for this limitation, they introduced a second branch composed of large-kernel convolutions~\cite{ding2022scaling}, which are known to emulate long-range dependencies without the computational overhead of attention modules. This branch emphasized global spatial interactions, enabling the model to better perceive large-scale structural degradation patterns across the entire face region. By aggregating the outputs from both local-detail-oriented and global-context-aware branches, their network effectively balanced sensitivity to local quality distortions with awareness of holistic structural integrity, which is especially critical in face IQA tasks where both local fidelity and global facial symmetry are important.

\paragraph{Training details.} They implemented their model using the PyTorch framework and trained it on a single NVIDIA RTX 4090 GPU. The network was optimized using the Adam optimizer with a learning rate of 2e-5 and a weight decay of 5e-4. The training process took approximately 2 hours. They adopted standard data preprocessing strategies including resizing, normalization, and random horizontal flipping. They trained for a total of 50 epochs.

\paragraph{Testing details.} The input images are resized to 224$\times$336 for testing.

\subsection{Knowledge Distillation for Improved Efficient MOdel(EMO) Image Quality Assessment (by 2077Agent)}
\begin{figure}[t]
  \centering
  \includegraphics[width=0.5\linewidth]{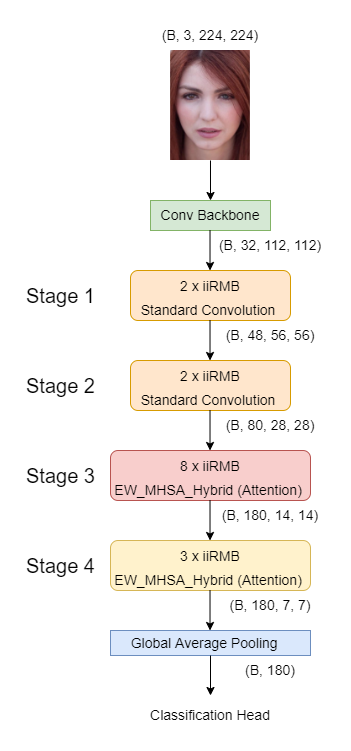}
  \caption{2077Agent.}
  \label{fig:2077agent}
\end{figure}

The \emph{2077Agent} Team first trained a high-capacity Efficient Model (EMO)~\cite{zhang2024emov2} on the Generic Face IQA (GFIQA) dataset to establish the teacher model. To meet the competition’s constraints on parameter count and computational complexity, they then designed and deployed a lightweight student model (shown in \cref{fig:2077agent}), enhancing its performance through a knowledge distillation strategy. Furthermore, they introduced optimized loss functions tailored to the evaluation metrics SRCC and PLCC to further improve the model’s predictive accuracy.

\paragraph{Training details.} Teacher and student networks share the same overall design—an initial convolutional layer that downsamples the input, followed by four hierarchical stages of improved inverted residual mobile blocks (iiRMB), and ending in a small regression head—but differ in depth, width, and resolution~\citep{zhang2024emov2}.

The teacher network processes $512\times512$ inputs and contains roughly 5 million parameters. After a three-layer stem (two $3\times3$ strided convolutions and a $1\times1$ projection), it has four stages with depths $\{3,3,9,3\}$. The embedding dimensions at each stage are $\{48,72,160,288\}$ with expansion ratios $\{2.0,3.0,4.0,4.0\}$. Early stages use BatchNorm+SiLU; later stages use LayerNorm+GELU. Regularization is minimal (\texttt{attn\_drop}=0, \texttt{drop}=0), with a drop-path rate of 0.05 and layer-scale initialized to $10^{-6}$. Finally, the feature map is global-average-pooled and passed through two BatchNorm+ReLU bottleneck layers before a sigmoid output~\cite{sandler2018mobilenetv2,zhang2023rethinking,zhuohang2024dynamic,zhang2024emov2}.

The student network scaled down to $1$ million parameters and $224\times224$ inputs, the student mirrors the teacher’s layout but with depths $\{2,2,8,3\}$ and embedding dimensions $\{32,48,80,180\}$. Expansion ratios are $\{2.0,2.5,3.0,3.5\}$, attention head dims $\{16,16,20,20\}$, and a $7\times7$ window. It uses the same BN+SiLU / LN+GELU scheme and a drop-path rate of 0.04036. Its regression head is identical to the teacher’s, ensuring comparable output capacity at a fraction of the size.

This training was conducted in two stages. First, the teacher model with approximately $5\times10^{6}$ parameters was trained for 100 epochs on a general face image quality assessment dataset. During training, they applied the following augmentations: \texttt{Normalize}; \texttt{RandCropOrResize} to randomly crop or resize images to $512\times512$; \texttt{RandHorizontalFlip}; and \texttt{ToTensor}. Optimization was performed with Adam (initial learning rate $1\times10^{-4}$, weight decay $1\times10^{-5}$) alongside a \texttt{StepLR} schedule that decayed the learning rate by a factor of 0.1 every 5 epochs. Upon completion, the teacher’s weights were frozen.

Next, they instantiated a student model under the same data loading and augmentation pipeline, and they performed widely used knowledge distillation method~\cite{deng2023improved, ouyang2024improving} using a \texttt{HybridLoss} ($\alpha = 0.75$) together with an MSE distillation loss (temperature = 4.0, weight = 0.5) to boost both SRCC and PLCC performance. The entire process ran on a single RTX 3090 GPU and took about 20 hours of training time.

\paragraph{Testing details.}
To avoid excessive computational load, they directly resize all test data to (224, 224) before feeding them into the network.

\subsection{LightHSPA: An Efficient and Lightweight Face Image Quality Assessment Network (by DERS)}

The \emph{DERS} Team proposed LightHSPA, a lightweight and efficient end-to-end convolutional neural network designed specifically for face image quality assessment (FIQA) under strict computational constraints. The architecture is composed of three main modules:
\begin{compactitem}  
\item Efficient Backbone: They employed a highly efficient feature extraction backbone inspired by MobileNetV2~\cite{sandler2018mobilenetv2}. It utilizes depthwise separable convolutions and inverted residual blocks to minimize parameter count and computational complexity (FLOPs). To further enhance feature representation with minimal overhead, they integrated a lightweight channel attention and an efficient spatial attention mechanism at the end of the backbone.
\item Lightweight HSPA Attention: To capture non-local dependencies and global facial context, they developed LightHSPA, a lightweight version of the High-order Self-attention with Projection Awareness (HSPA) mechanism. This module uses depthwise separable convolutions and adaptive pooling to significantly reduce the computational cost of the self-attention operation, making it suitable for an efficient model. This module was used during experimentation to explore performance trade-offs.
\item Compact Quality Head: A compact regression head predicts the final quality score. It uses multi-scale global pooling (average and max) to create a robust feature vector from the final feature map. This vector is then processed by a small multi-layer perceptron (MLP) with dropout and batch normalization to produce a single scalar quality score.
\end{compactitem}

The entire model is designed to be configurable (e.g., via width multiplier) to balance the trade-off between performance (SRCC/PLCC) and efficiency (parameters/FLOPs).

\paragraph{Training details.} 
Their model was implemented in PyTorch and trained on a single NVIDIA 4090 GPU for 200 epochs, taking approximately 9 hours. They used the Adam optimizer with an initial learning rate of $2\times10^{-3}$ and a weight decay of $1\times10^{-4}$, employing a step learning rate scheduler that decayed the rate by a factor of 0.1 every 5 epochs. Only the official 30,000 in-the-wild face images from the competition were used for training. Their training strategy involved minimizing Mean Squared Error (MSE) loss and applying data augmentation, including random cropping to $224\times224$ patches and random horizontal flipping.

\paragraph{Testing details.}
Their model takes a single face image as input and outputs a single perceptual quality score. No test-time augmentation (TTA) or other post-processing steps were used. They ran the model on the original image resolutions as provided in the test set.
\fi
\paragraph{Acknowledgements.} We thank Snap for sponsoring this challenge. We would also like to thank the VQualA 2025 sponsors: Snap, Intsig, and Taobao and Tmall Group.

{
    \small
    \bibliographystyle{ieeenat_fullname}
    \bibliography{main}
}
\ifarxiv
\appendix
\section{Teams and Affiliations}
\label{sec:team}
\subsection*{VQualA 2025 FIQA Track Organizers}
\noindent\textbf{Members:} Sizhuo Ma\tss{1} (\email{sma@snap.com}), Wei-Ting Chen\tss{2} (\email{weitingchen@microsoft.com}),  Qiang Gao\tss{1} (\email{qgao@snap.com}), Jian Wang\tss{1}(\email{jwang4@snap.com}) and Chris Wei Zhou\tss{3} (\email{zhouw26@cardiff.ac.uk})\\
\textbf{Affiliations:}\\
\tss{1}Snap Inc.\\
\tss{2}Microsoft\\
\tss{3}Cardiff University

\subsection*{ECNU-SJTU VQA Team}
\noindent\textbf{Proposed Method:} Efficient Face Image Quality Assessment via Self-training and Knowledge Distillation\\
\textbf{Members:} Wei Sun\tss{1} (\email{sunguwei@gmail.com}), Weixia Zhang\tss{2} (\email{zwx8981@sjtu.edu.cn}),  Linhan Cao\tss{2} (\email{caolinhan@sjtu.edu.cn}), Jun Jia\tss{2}(\email{jiajun0302@sjtu.edu.cn}), Xiangyang Zhu\tss{3} (\email{zhuxiangyang@pjlab.org.cn}), Dandan Zhu\tss{1} (\email{ddzhu@mail.ecnu.edu.cn}), Xiongkuo Min\tss{2} (\email{minxiongkuo@sjtu.edu.cn}) and Guangtao Zhai\tss{2} (\email{zhaiguangtao@sjtu.edu.cn})\\
\textbf{Affiliations:}\\
\tss{1}East China Normal University\\
\tss{2}Shanghai Jiao Tong University\\
\tss{3}Shanghai Artificial Intelligence Laboratory

\subsection*{MediaForensics}
\noindent\textbf{Proposed Method:} MSPT: A Lightweight Face Image Quality Assessment Method with Multi-Stage Progressive Training\\
\textbf{Members:} Baoying Chen\tss{1} (\email{1900271059@email.szu.edu.cn}), Xiongwei Xiao\tss{2} (\email{xiongweixiaoxxw@gmail.com}) and Jishen Zeng\tss{1} (\email{jishen.zjs@alibaba-inc.com})\\
\textbf{Affiliations:}\\
\tss{1}Alibaba Group\\
\tss{2}The Hong Kong Polytechnic University

\subsection*{Next}
\noindent\textbf{Proposed Method:} Towards Robust No-Reference Image Quality Assessment via Prompt-Aware Alignment and Multi-Level Distillation\\
\textbf{Members:} Wei Wu\tss{1} (\email{wuwei.lorenzo@gmail.com}), Tiexuan Lou\tss{2} (\email{xuanxuanqaq@gmail.com}), Yuchen Tan\tss{2} (\email{12334068@zju.edu.cn}), Chunyi Song\tss{2} (\email{cysong@zju.edu.cn}) and Zhiwei Xu\tss{2} (\email{xuzw@zju.edu.cn})\\
\textbf{Affiliations:}\\
\tss{1}Donghai Laboratory\\
\tss{2}Ocean College, Zhejiang University

\subsection*{ATHENAFace}
\noindent\textbf{Proposed Method:} Face Image Quality Assessment via Lightweight Ensemble Learning and Correlation-Driven Optimization\\
\textbf{Members:} MohammadAli Hamidi\tss{1} (\email{mohammadali.hamidi@unica.it}) and Hadi Amirpour\tss{2} (\email{hadi.amirpour@aau.at})\\
\textbf{Affiliations:}\\
\tss{1}University of Cagliari\\
\tss{2}University of Klagenfurt

\subsection*{NJUPT-IQA-Group}
\noindent\textbf{Proposed Method:} Lightweight Spatial-Frequency Fusion Network for Blind Face Image Quality Assessment\\
\textbf{Members:} Mingyin Bai (\email{1223014043@njupt.edu.cn}), Jiawang Du (\email{1223013838@njupt.edu.cn}), Zhenyu Jiang (\email{1224014031@njupt.edu.cn}), Zilong Lu (\email{1224014206@njupt.edu.cn}), Ziguan Cui (\email{cuizg@njupt.edu.cn}) and Zongliang Gan (\email{ganzl@njupt.edu.cn})\\
\textbf{Affiliations:}\\
\tss{1}School of Communications and Information Engineering, Nanjing University of Posts and Telecommunications

\subsection*{ECNU VIS Lab}
\noindent\textbf{Proposed Method:} MobileNetV4 in FIQA\\
\textbf{Members:} Xinpeng Li\tss{1} (\email{51275901118@stu.ecnu.edu.cn}), Shiqi Jiang\tss{1} (\email{52265901032@stu.ecnu.edu.cn}), Chenhui Li\tss{1} (\email{chli@cs.ecnu.edu.cn}) and Changbo Wang\tss{1} (\email{cbwang@dase.ecnu.edu.cn})\\
\textbf{Affiliations:}\\
\tss{1}East China Normal University

\subsection*{JNU620}
\noindent\textbf{Proposed Method:} Progressive Learning Strategy for Face Image Quality Assessment\\
\textbf{Members:} Weijun Yuan\tss{1} (\email{yweijun@stu2022.jnu.edu.cn}), Zhan Li\tss{1} (\email{lizhan@jnu.edu.cn}), Yihang Chen\tss{1} (\email{ehang@stu.jnu.edu.cn}), Yifan Deng\tss{1} (\email{dyf010408@stu.jnu.edu.cn}), Ruting Deng\tss{1} (\email{routine@stu2022.jnu.edu.cn}), Zhanglu Chen\tss{1} (\email{czhanglu@stu2024.jnu.edu.cn}), Boyang Yao\tss{1} (\email{yaoboy@stu.jnu.edu.cn}), Shuling Zheng\tss{1} (\email{3440989938@qq.com}), Feng Zhang\tss{1} (\email{1569259893@qq.com}) and Zhiheng Fu\tss{1} (\email{2557502986@qq.com})\\
\textbf{Affiliations:}\\
\tss{1}Jinan University

\subsection*{ISeeCV}
\noindent\textbf{Proposed Method:} RankCORE: Ranking-Aware Correlation Optimized Regression Estimator\\
\textbf{Members:} Abhishek Joshi\tss{1} (\email{abhishek.j@aftershoot.com}) and Aman Agarwal\tss{1} (\email{aman.a@aftershoot.com})\\
\textbf{Affiliations:}\\
\tss{1}Aftershoot

\subsection*{RegNet}
\noindent\textbf{Proposed Method:} Face Image Quality Assessment in the Wild with RegNet\\
\textbf{Members:} Rakhil Immidisetti\tss{1} (\email{samrakhil@gmail.com}) and Ajay Narasimha Mopidevi\tss{2} (\email{ajaymopidevi@gmail.com})\\
\textbf{Affiliations:}\\
\tss{1}Blue River Technology\\
\tss{2}Lendbuzz

\subsection*{Conquerit}
\noindent\textbf{Proposed Method:} MobileNetV3-Based FIQA \\
\textbf{Members:} Vishwajeet Shukla\tss{1} (\email{vishwajeet993511@gmail.com})\\
\textbf{Affiliations:}\\
\tss{1}Adobe Systems

\subsection*{BIT\_ssvgg}
\noindent\textbf{Proposed Method:} Dual-Branch Network with Local and Global Perception for Face Image Quality Assessment\\
\textbf{Members:} Hao Yang\tss{1} (\email{3120235187@bit.edu.cn}), Ruikun Zhang\tss{1} (\email{ruikun.zhang@bit.edu.cn}) and Liyuan Pan\tss{1} (\email{liyuan.pan@bit.edu.cn})\\
\textbf{Affiliations:}\\
\tss{1}School of Computer Science \& Technology, Beijing Institute of Technology

\subsection*{2077Agent}
\noindent\textbf{Proposed Method:} Knowledge Distillation for Improved Efficient MOdel(EMO) Image Quality Assessment\\
\textbf{Members:} Kaixin Deng\tss{1} (\email{kaixin.deng.t0@elms.hokudai.ac.jp}), Hang Ouyang\tss{2} (\email{1246276321@qq.com}), Fan Yang\tss{2} (\email{173001344@qq.com}), Zhizun Luo\tss{2} (\email{1151507490@qq.com}) and Zhuohang Shi\tss{3} (\email{shi\_zhuo\_hang@yeah.net})\\
\textbf{Affiliations:}\\
\tss{1}Hokkaido University\\
\tss{2}Chengdu University of Technology\\
\tss{3}Hebei University of Technology

\subsection*{DERS}
\noindent\textbf{Proposed Method:} LightHSPA: An Efficient and Lightweight Face Image Quality Assessment Network\\
\textbf{Members:} Songning Lai\tss{1} (\email{songninglai@hkust-gz.edu.cn}), Weilin Ruan\tss{1} (\email{wruan792@connect.hkust-gz.edu.cn}) and Yutao Yue\tss{1} (\email{yutaoyue@hkust-gz.edu.cn})\\
\textbf{Affiliations:}\\
\tss{1}The Hong Kong University of Science and Technology (Guangzhou)
\section{Details about RankCORE} 
\label{sec:rankcore}
This appendix provides details about the RankCORE model developed by the ISeeCV Team.

\paragraph{Self-Supervised Adaptive Ranking (SSAR).}
Traditional Perceptual Quality Assessment (PQA) methods~\cite{mittal2012no, kang2014blind} require large datasets of images annotated with scalar quality scores—a process that is \emph{expensive}, \emph{subjective}, and \emph{time-consuming}. \emph{Self-supervised learning} (SSL)~\cite{chen2020simple} sidesteps this requirement by exploiting \emph{relative relationships} between samples rather than absolute labels.

They observed that common image degradations—such as Gaussian blur, additive noise, and interpolation artifacts—\emph{monotonically degrade} perceptual quality: if I is an original face, then its transformed version $I'=T_s(I)$ always satisfies
\[
  q(I') < q(I),
\]
where $T_s$
  is a transformation at severity  $s\in[0,1]$. They leveraged this inherent ordering to train a PQA network using \emph{margin-based ranking loss}, enabling the model to \emph{learn absolute prediction} of quality without ever seeing ground-truth scores.

For each pair $(I,I')$, with predicted scores $\hat q(I)$, $\hat q(I')$, and severity $s$, they defined margin
\[
m(s) = \lambda\,s, \hspace{5mm}  \lambda > 0
\]
and optimized:
\[
\mathcal{L}_{\mathrm{adaptive}}(I,I';s) = \max\bigl(0,\,-(\hat q(I)-\hat q(I')) + m(s)\bigr).
\]
Here, larger severities enforce larger margins, compelling the network to respect greater quality gaps when distortions are harsher.

\paragraph{Score‑Stratified Uniform Sampler (SSUS).}
They introduced SSUS, which uniformly selects score-strata and then uniformly samples within each stratum to guarantee balanced coverage of the full score distribution. This ensured that rare or underrepresented score regions were seen as often as dense ones, reducing bias and improving model robustness. It also promoted gradient diversity, stabilized training, and could be extended by weighting specific bins. 

\paragraph{CoReFace: Correlation-Robust Face Quality Estimation.}
To penalize both local prediction errors and global ranking mistakes, they adopted a composite Wing-PLCC loss. It is well known that MSE is not sensitive to outliers, while L1 loss doesn't penalize mid-range errors in regression. WingLoss~\cite{feng2018wing} brings the best of both worlds. To the best of their knowledge, WingLoss has not been fundamentally explored for face perceptual quality assessment.
WingLoss~\cite{feng2018wing} ($\omega$ = 0.03, $\epsilon$ = 2)  gives logarithmic sensitivity around small absolute errors while retaining linear behavior for larger discrepancies, helping the network focus on hard-to-distinguish quality levels, as shown in \cref{eq:wingloss}.
Additionally, they leveraged the Pearson Linear Correlation Coefficient (PLCC) metric as a differentiable loss function. They augmented this with a global PLCC term to align predictions with the mean-opinion-score distribution, as defined in \cref{eq:plcc}. 
    \begin{equation}
    \label{eq:wingloss}
WingLoss =
\begin{cases}
w \ln\left(1 + \frac{|x|}{\epsilon} \right), & \text{if } |x| < w \\
|x| - C, & \text{otherwise}
\end{cases}
\end{equation}

    \begin{equation}
    \label{eq:plcc}
    PLCCLoss(y, \hat{y}) = 1 - \frac{\sum_{i=1}^{n} (y_i - \bar{y})(\hat{y}_i - \bar{\hat{y}})}
{\sqrt{\sum_{i=1}^{n} (y_i - \bar{y})^2} \sqrt{\sum_{i=1}^{n} (\hat{y}_i - \bar{\hat{y}})^2}}
    \end{equation}

The final CoReFaceLoss is defined as the weighted sum of \cref{eq:wingloss} and \cref{eq:plcc}:
    \begin{equation}
   CoReFace Loss = \alpha \cdot WingLoss + \beta \cdot PLCCLoss  \end{equation} 




\fi

\end{document}